\documentclass{article}

\usepackage{microtype}
\usepackage{graphicx}
\usepackage{subfigure}
\usepackage{booktabs} 
\usepackage{amsfonts}
\usepackage{amsmath}
\usepackage{color}

\usepackage{hyperref}

 \usepackage[accepted]{icml2021}

\icmltitlerunning{Self-Supervised Noisy Label Learning for Source-Free Unsupervised Domain Adaptation}

\begin{document}

\twocolumn[
\icmltitle{Self-Supervised Noisy Label Learning for \\
Source-Free Unsupervised Domain Adaptation}



\icmlsetsymbol{equal}{*}

\begin{icmlauthorlist}
\icmlauthor{Weijie Chen}{equal,zju,hik}
\icmlauthor{Luojun Lin}{equal,fzu}
\icmlauthor{Shicai Yang}{hik}
\icmlauthor{Di Xie}{hik}
\icmlauthor{Shiliang Pu}{hik}
\icmlauthor{Yueting Zhuang}{zju}
\icmlauthor{Wenqi Ren}{hik}\\
\icmlauthor{chenweijie5@hikvision.com}{}
\icmlauthor{linluojun2009@126.com}{}\\
\icmlauthor{\{yangshicai,xiedi,pushiliang.hri,renwenqi\}@hikvision.com}{}
\icmlauthor{yzhuang@zju.edu.cn}{}
\end{icmlauthorlist}

\icmlaffiliation{zju}{Zhejiang University, Hangzhou, China}
\icmlaffiliation{fzu}{Fuzhou University, Fuzhou, China}
\icmlaffiliation{hik}{Hikvision Research Institute, Hangzhou, China}

\icmlcorrespondingauthor{Yueting Zhuang}{yzhuang@zju.edu.cn}

\icmlkeywords{Unsupervised Domain Adaptation, Noisy Label Learning, Self-Supervised Learning}

\vskip 0.3in
]



\printAffiliationsAndNotice{\icmlEqualContribution} 

\begin{abstract}
It is a strong prerequisite to access source data freely in many existing unsupervised domain adaptation approaches. However, source data is agnostic in many practical scenarios due to the constraints of expensive data transmission and data privacy protection. Usually, the given source domain pre-trained model is expected to optimize with only unlabeled target data, which is termed as source-free unsupervised domain adaptation. In this paper, we solve this problem from the perspective of noisy label learning, since the given pre-trained model can pre-generate noisy label for unlabeled target data via directly network inference. Under this problem modeling, incorporating self-supervised learning, we propose a novel \emph{Self-Supervised Noisy Label Learning} method, which can effectively fine-tune the pre-trained model with pre-generated label as well as self-generated label on the fly. Extensive experiments had been conducted to validate its effectiveness. Our method can easily achieve state-of-the-art results and surpass other methods by a very large margin. Code will be released. 
\end{abstract}

\section{Introduction}
\begin{figure}[t]
\vskip 0.2in
\begin{center}
\centerline{\includegraphics[width=1.0\columnwidth]{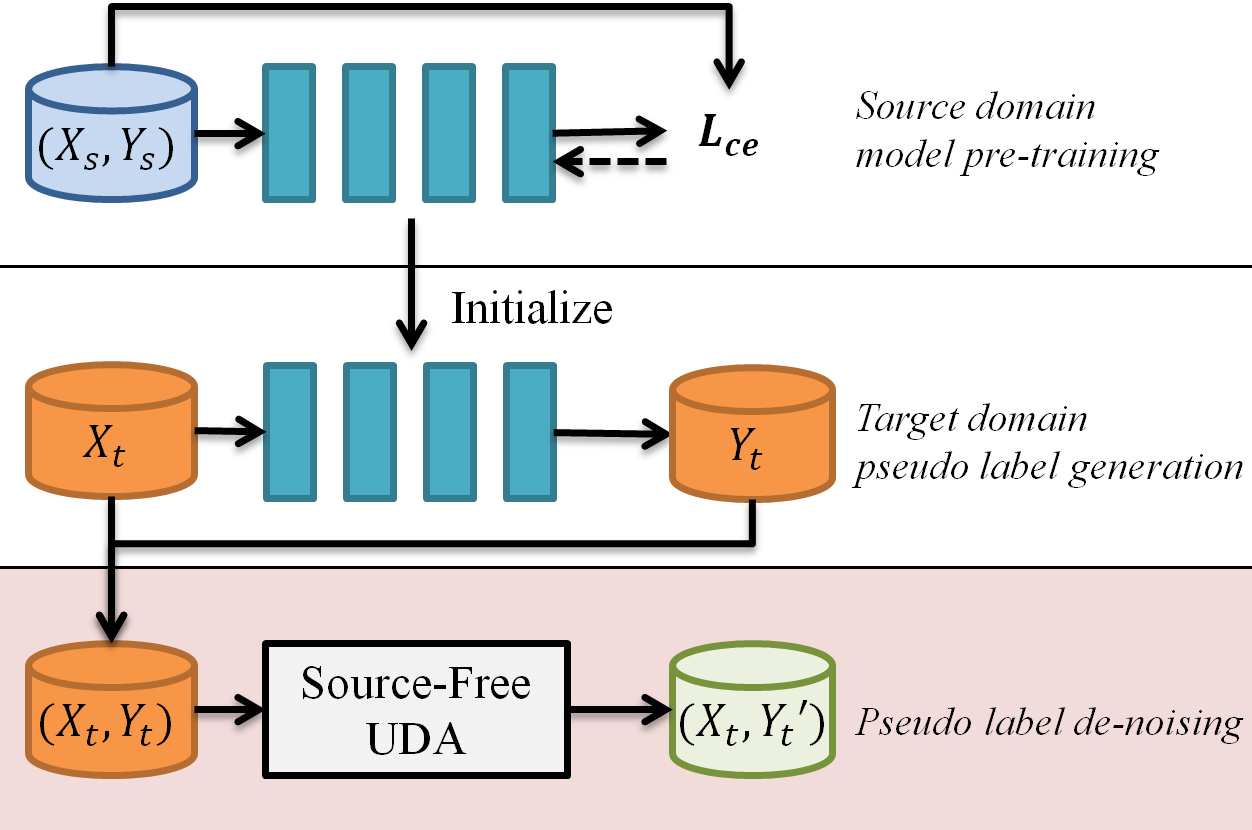}}
\caption{Source-free UDA can be viewed as pseudo label de-noising. Here $(\mathcal{X}_s,\mathcal{Y}_s)$ denote the source data and the annotated label. $\mathcal{X}_t$ is the unlabeled target data. The pre-generated pseudo label $\mathcal{Y}_t$ is denoised into a cleaner one $\mathcal{Y}_t'$ after source-free UDA.}
\label{SFUDA}
\end{center}
\vskip -0.2in
\end{figure}


In practical applications, a deep model trained on source domain is usually deployed in edge devices to test unlabeled images from unknown target domain. The data distribution of target domain is rather different from source domain due to agnostic domain shift, such as diverse illumination, complex weather, etc. This is the main factor of model performance degradation in real-world scenarios. Recently, there are more and more researchers delving into unsupervised domain adaptation (UDA) to address this problem.


Most of previous UDA methods aim to align the labeled source data and unlabeled target data in a common representation space, so that the classifier trained on source domain can be well generalized to target domain~\cite{MMD, Deepcoral, AssocDA}. These vanilla UDA methods always assume that source data is accessible and thus can be used with target data for domain transferring. However, it is ill-suited in some practical applications, $e.g.$, source data is inaccessible and only the model pre-trained on source domain is available due to the expensive data transmission and data privacy protection. Such situation is termed as source-free UDA, where only unlabeled target data is provided for model optimization~\cite{MADA, SHOT-1M, SFOD}. Note that we only discuss image classification task in this paper.

How to solve source-free UDA? Although the pre-trained model performs not so well on target domain, it still contains informative cues of the task. Naturally, it can be exploited to pre-generate pseudo labels for target data via network inference. Inevitably, the pre-generated labels are not exactly correct, where the false-labeled ones can be viewed as noisy labels. In this way, as shown in Fig.\ref{SFUDA}, source-free UDA can also be regarded as another form of noisy label learning. From this viewpoint, we propose a simple yet effective approach named Self-Supervised Noisy Label Learning (SSNLL) to address this problem, which is strongly inspired by unsupervised image classification and noisy label learning.


First of all, we walk from Unsupervised Image Classification (UIC) method \cite{UIC}, which is an unsupervised technique to train the classification network by self-generating pseudo label. Similar to other self-supervised learning methods \cite{MOCO, SimCLR, DeepCluster}, a critical step is to avoid model collapse of classifying all images into one category. Besides, the classification results achieved in a totally unsupervised way cannot be directly used in downstream tasks. Hence, we consider to incorporate noisy label learning to eliminate these problems.


Since the pre-generated noisy label contains informative cues of the task, it can regularize UIC towards a task-specific optimization direction. Considering what if we can split target data into a true-labeled part and a false-labeled part? The true-labeled part trained with pre-generated labels can regularize the false-labeled part trained with self-generated labels. Only when the self-generated label in false-labeled part matches to the pre-generated pseudo label in true-labeled part with consistent semantic information can it achieve optimal solution.

To achieve this objective, inspired by the small-loss trick used in noisy label learning \cite{mentornet,meta-weight-net,Co-teaching}, we split the target data $\mathcal{X}_{t}$ into a cleaner part $\mathcal{X}_{cl}$ with smaller loss and a noisier part $\mathcal{X}_{no}$ with greater loss with respect to the pre-generated label. To avoid the aforementioned model collapse problem, we further develop it into a label-wise dataset splitting method, which ensures no empty classes in the cleaner part $\mathcal{X}_{cl}$. After that, we sample images from $\mathcal{X}_{cl}$ and $\mathcal{X}_{no}$ uniformly to train the network with pre-generated label and self-generated label, respectively. As the training goes, the loss with respect to the fixed pre-generated label will get smaller in true-labeled samples and get larger in false-labeled samples. To fully exploit this positive feedback, the dataset splitting operation and network training operation are alternated epoch by epoch so as to progressively boost the performance.


Actually, UIC and noisy label learning in our approach are promoted by each other. The former one can help refine the pre-generated noisy label, whilst the latter one can regularize self-generated label and prevent UIC from model collapsing and class mismatching. Besides, in order to initially reduce the noise ratio, we also introduce two label denoising tricks during the process of pre-generating pseudo labels, including Adaptive Batch Normalization (AdaBN) \cite{Li2016Revisiting} and Deep Transfer Clustering (DTC) \cite{DTC}.

Our method can well-solve the source-free UDA problem. Extensive experiments had been carried out on several popular UDA benchmarks, which show our method can easily achieve state-of-the-art results on these benchmarks. It surpasses other methods even the source data-based ones by a very large margin. For instance, on VisDA-C\cite{Peng2017VisDA}, one of the most challenging datasets in UDA, our method can achieve 85.8\% accuracy and surpass the second place more than 3\% accuracy.

\section{Related Work}
\textbf{Unsupervised Domain Adaptation.} Vanilla unsupervised domain adaptation means that the model is trained on labeled source data and unlabeled target data jointly, and then the knowledge of source domain can be transferred to target domain adaptively without any extra annotations. In early stage, some discrepancy-based methods are proposed to minimize the well-defined distance loss functions of two domains~\cite{MMD, Deepcoral, AssocDA}. Specifically, adversarial learning is also used in UDA, where domain gap is minimized in feature-level by gradient reversal layer~\cite{GRL}, generative adversarial networks~\cite{CoGAN, PixelDA, CYCADA}, or hybrid methods~\cite{MCD, DIRT-T}. 

However, some special situations are occurred that source data is inaccessible and only a source domain pre-trained model is available due to the constraints of data transmission and data privacy protection. This kind of source-free UDA problem can be considered as unsupervised target domain learning, which is more difficult to resolve than vanilla UDA because of less supervision information. Model adaptation~\cite{MADA} employs source model as an auxiliary classifier of GAN to synthesize target-style false images, and these false images are used to fine-tune source classifier. SHOT-1M~\cite{SHOT-1M} utilizes DeepCluster~\cite{DeepCluster} to generate pseudo labels fine-tuning source model. Different from the above methods, we treat source-free UDA as a noisy label learning task, which eases the problem and achieves superior performance. 

\textbf{Noisy Label Learning.} A phenomenon called memorization effects is discovered in noisy label learning that deep networks are prone to fit easy (clean) samples, and gradually over-fit hard (noisy) samples~\cite{memorization}. Hence, sample selection is proposed to solve noisy label learning problem (NLL) by selecting clean samples according to some rules. SELF~\cite{SELFE} uses the prediction consistency between the ensemble network outputs and labels to filter out noisy-labeled samples. Besides, researchers adopt another sample selection criterion, namely \emph{small-loss} trick, which regards samples with small-loss as clean samples and only back-propagates such samples to update network parameters. MentorNet~\cite{mentornet} introduces curriculum learning into NLL, by using mentor net to provide curriculum about how to select small-loss samples for student net. Meta-weight-net~\cite{meta-weight-net} constructs a meta-net to assign weights for each sample, where small-loss samples tend to be assigned larger weights and vice versa. Co-teaching~\cite{Co-teaching} / Co-teaching+~\cite{Co-teaching+} trains two networks simultaneously, and each network selects a certain number of small-loss samples from the same batch. Then, each network is updated by back-propagating the samples selected by its peer network. In this paper, we adopt small-loss trick in source-free UDA task by simply dividing target dataset into a cleaner subset and a noisier subset. The difference with other works is that we preserve rather than drop the noisy subset. Furthermore, we develop a label-wise dataset splitting method to avoid model collapsing.  

\textbf{Self-Supervised Learning.} Without any supervision from human-annotated labels, most of self-supervised learning works (SSL) facilitate contrastive learning method that forces model to distinguish an anchor with its positive images and negative images~\cite{MOCO, SimCLR}. Different from contrastive learning, DeepCluster~\cite{DeepCluster} iterates between clustering the features by $k$-means and updating the network by predicting the cluster assignments as pseudo labels in discriminative loss. To avoid the memory-cost and inefficiency brought by storing all sample features in DeepCluster, an easier method, namely Unsupervised Image Classification (UIC), is proposed to employ softmax layer rather than $k$-means to generate pseudo labels, where softmax is regarded as an implicit clustering function~\cite{UIC}. It iterates between pseudo label generation and network optimization per epoch. 
In this paper, we follow UIC that use self-generated pseudo labels to refine noisy subset, whilst using pre-generated noisy labels to regularize and prevent UIC from model collapsing.

\section{Method}

\subsection{Problem Setup} 
Source-free UDA can be regarded as a two-stage framework. Firstly, source data $\mathcal{X}_s$ is collected and annotated as $\mathcal{Y}_s$ to train the model $f$. For image classification task, cross-entropy loss $\mathcal{L}_{ce}$ is usually selected to train $f$:
\begin{equation}
\min\limits_{f} \mathcal{L}_{ce}(f | \mathcal{X}_s, \mathcal{Y}_s)
\end{equation}
Secondly, given the pre-trained model $f$, it is a great challenge to optimize $f$ with unlabeled target data $\mathcal{X}_t$ merely. In this paper, it is formulated as a noisy label learning task. We exploit $f$ to pre-generate noisy label $\mathcal{Y}_t$ for target data $\mathcal{X}_t$:
\begin{equation}
\mathcal{Y}_t=\arg\max f(\mathcal{X}_t)
\label{e2}
\end{equation}
In this way, the second stage of source-free UDA is exactly turned into how to fine-tune $f$ with pre-generated noisy label $(\mathcal{X}_t, \mathcal{Y}_t)$:
\begin{equation}
\min\limits_{f} \mathcal{L}(f | \mathcal{X}_t, \mathcal{Y}_t)
\end{equation}
Different from traditional noisy label learning task, in source-free UDA, $f$ is pre-trained on labeled source data at first and the noisy label of target data is naturally generated rather than manually or randomly setting. 

\subsection{Label Denoising Preprocessing}
\begin{figure}[tb]
\vskip 0.2in
\begin{center}
\centerline{\includegraphics[width=0.85\columnwidth]{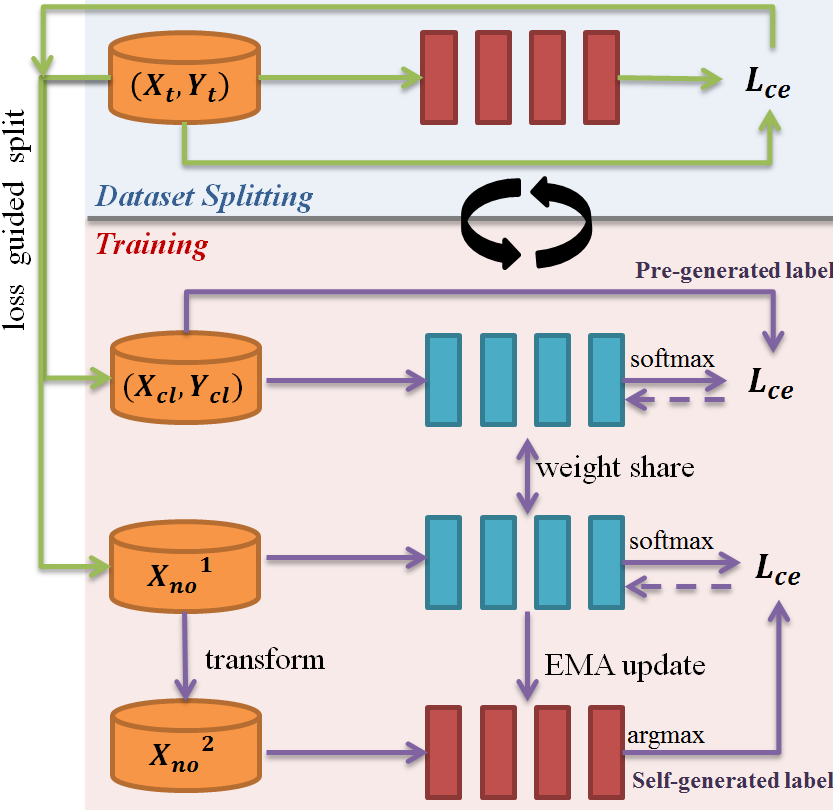}}
\caption{The framework of \emph{Self-Supervised Noisy Label Learning} for source-free unsupervised domain adaptation. Two steps are alternated epoch by epoch: 1) Target data $(\mathcal{X}_t,\mathcal{Y}_t)$ is splitted into a cleaner part $(\mathcal{X}_{cl},\mathcal{Y}_{cl})$ and a noisier part $\mathcal{X}_{no}$ according to a small-loss trick. 2) During training, the former part is optimized with pre-generated label while the latter part is optimized with self-generated label on the fly.}
\label{SSNLL}
\end{center}
\vskip -0.2in
\end{figure}
Can we rectify the pre-generated noisy label $\mathcal{Y}_t$ before fine-tuning $f$? A lower noise ratio can naturally bring benefit to the following noisy label learning process. Here we adopt two label denoising tricks:

\textbf{Adaptive Batch Normalization.} As illustrated in AdaBN \cite{Li2016Revisiting}, the population statistics ($i.e.$, mean $\mu$ and variance $\sigma$) of Batch Normalization (BN) layers encode the domain-specific information. When the data distribution is changed from source domain to target domain, the population statistics of BN layer should be recalculated, such that the feature representation will be transferred adaptively, which benefits to achieve cleaner pseudo label $\mathcal{Y}_t$ on target domain. It is prone to implement AdaBN by feeding batches of target data into the pre-trained model $f$ to calculate batch statistics, and the population statistics is updated by a momentum-based moving average of batch statistics:
\begin{equation}
\{\mu,\sigma\}=\lambda\{\mu,\sigma\}+(1-\lambda)\{\mu,\sigma\}_{batch}
\label{e4}
\end{equation}
where $\{\mu,\sigma\}_{batch}$ denotes the statistics of current batch on target domain, and $\{\mu,\sigma\}$ is the population statistics initialized by the statistics from source domain. After updated on the whole target data, the population statistics are utilized to pre-generate pseudo labels via network inference. 

\textbf{Deep Transfer Clustering.} Under the assumption that the distribution of features is consistent to the corresponding labels \cite{Rui2005Survey,2008Class}, we use deep transfer clustering (DTC) \cite{DTC} to denoise pre-generated pseudo labels $\mathcal{Y}_t$. Through this operation, the samples in the same cluster are forced to share the same labels.

Specifically, we extract the features of target data for $k$-means clustering, where the feature extractor $e$ is separated from the pre-trained source model $f$:
\begin{equation}
\min_{\mathcal{M}\in \mathbb{R}^{d\times k}} \frac{1}{N} \sum\limits_{n=1}^{N}\min_{c_n\in\{0,1\}^k}||e(x_n)-\mathcal{M}c_n||_{2}^{2} \,\, {\rm s.t.}\,\, c_n^T\textbf{1}_k=1
\label{e5}
\end{equation}
where $x_n$ and $c_n$ denote the $n$-th target sample and its cluster assignment, respectively. N is the sample number. $\mathcal{M}\in \mathbb{R}^{d\times k}$ is the cluster centroid matrix with size of $d\times k$, where $d$ is the feature dimension and $k$ is the cluster number. To ensure the sample consistency within each cluster, over-clustering is necessary. We set $k$ as $10\times$ of the class number in the corresponding tasks. After clustering, the pseudo labels are denoised through aggregating the probability distributions within each cluster:
\begin{equation}
\widehat{p_n}=\frac{1}{|\Omega|} \sum\limits_{i\in\Omega}p_i\,\,{\rm where}\,\,\Omega=\{j \,\,|\,\, c_j=c_n\}
\label{e6}
\end{equation}
\begin{equation}
\widehat{y_n}=\arg\max \,\, \widehat{p_n}
\label{e7}
\end{equation}
where $p_n$ is the classification probability of $x_n$ (the output of softmax) and $\widehat{p_n}$ is the refined one after aggregation. We rebuild a cleaner label set $\widehat{\mathcal{Y}_t}=\{\widehat{y_n}\}$ to conduct the following self-supervised noisy label learning process.

\subsection{Self-Supervised Noisy Label Learning}
\begin{figure}[tb]
\vskip 0.2in
\begin{center}
\centerline{\includegraphics[width=0.75\columnwidth]{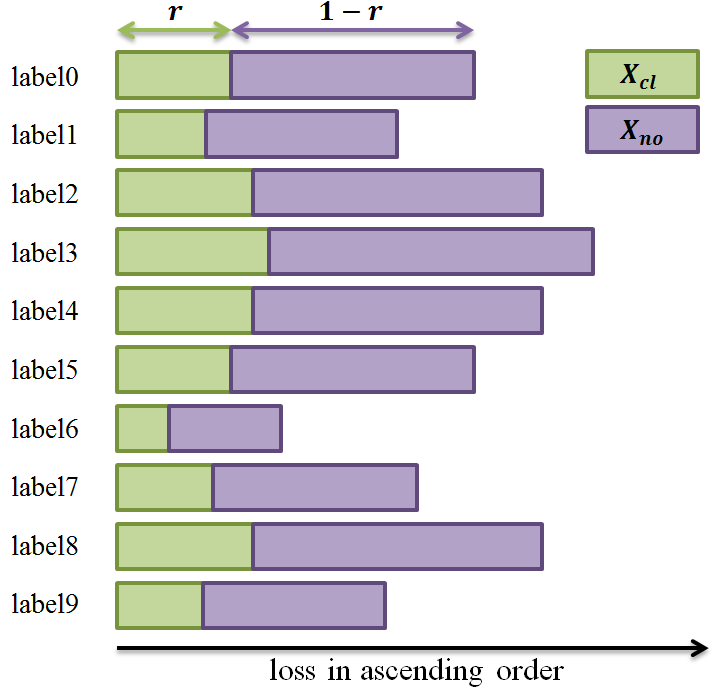}}
\caption{Label-wise dataset splitting method. According to small-loss trick, the target data is splitted into a cleaner part $\mathcal{X}_{cl}$ and a noisier part $\mathcal{X}_{no}$ with a hyper-parameter split ratio $r$. To prevent empty classes in $\mathcal{X}_{cl}$, the dataset is splitted in a label-wise way.}
\label{DatasetSplit}
\end{center}
\vskip -0.2in
\end{figure}
Before introducing our approach, we first review the recent development of unsupervised image classification (UIC) \cite{UIC}. It has become a mature technique nowadays, which generates pseudo label to drive image classification in a self-supervised way:
\begin{equation}
\widetilde{y_n}=\arg\max f(x_n^2)
\label{e8}
\end{equation}
\begin{equation}
\min\limits_{f}\mathcal{L}_{ce}(f(x_n^1), \widetilde{y_n}))
\label{e9}
\end{equation}
where $x_n^1$ and $x_n^2$ are two different random transformations of $x_n$. $\widetilde{y_n}$ is the self-generated pseudo label. For simplicity, Eqn.\ref{e8} and \ref{e9} can be combined as:
\begin{equation}
\min\limits_{f}\mathcal{L}_{ce}(f(x_n^1), \arg\max f(x_n^2)))
\label{e10}
\end{equation}
An important training trick in this learning process is to prevent model collapsing that all input samples tend to be classified into one category. In \cite{UIC}, they prevent model collapsing via avoiding empty class and ensuring balanced sampling. Besides, UIC is not directly developed for downstream tasks. In source-free UDA, the pre-generated noisy label is given to incorporate with UIC to learn task-specific image classification.

Our idea is intuitive, which splits the entire dataset into a cleaner part $\mathcal{X}_{cl}$ and a noisier part $\mathcal{X}_{no}$, and trains these two parts with pre-generated and self-generated label, respectively. According a small-loss trick, which is a popular criterion to select true-labeled samples during noisy label learning, the samples with small loss tend to be true-labeled samples. Therefore, these two parts can be splitted guided by the loss with respect to the pre-generated label:
\begin{equation}
\{(\mathcal{X}_{cl}, \widehat{\mathcal{Y}_{cl})}\},\{\mathcal{X}_{no}\} \leftarrow \mathcal{L}_{ce}(f(\mathcal{X}_t), \widehat{\mathcal{Y}_t})
\label{e12}
\end{equation}
To make sure there is no empty classes in $\{(\mathcal{X}_{cl}, \widehat{\mathcal{Y}_{cl}})\}$ so as to avoid the aforementioned model collapse, we propose a label-wise dataset splitting method, which divides the dataset into several groups based on pseudo label and then sort the loss within each group. Given a split ratio $r$ as a hyper-parameter, the samples in each group can be splitted into two parts to construct $\mathcal{X}_{cl}$ and $\mathcal{X}_{no}$ as shown in Fig.\ref{DatasetSplit}.

After dataset splitting, $\mathcal{X}_{cl}$ is directly trained with $\widehat{\mathcal{Y}_{cl}}$ pre-generated by the pre-trained model, while $\mathcal{X}_{no}$ is trained with the self-generated label on the fly as shown in Eqn.\ref{e10}. The training of the former one actually regularizes the latter one to learn task-specific information. To balance these two terms, $\mathcal{X}_{cl}$ and $\mathcal{X}_{no}$ are sampled and packed into one batch uniformly for training in each iteration. The training objective of this process can be formulated as follows:
\begin{equation}
\min\limits_{f}\mathcal{L}_{ce}(f(\mathcal{X}_{cl}), \widehat{\mathcal{Y}_{cl}})+\mathcal{L}_{ce}(f(\mathcal{X}_{no}^1), \arg\max f(\mathcal{X}_{no}^2))
\label{e11}
\end{equation}

Eqn.\ref{e12} and Eqn.\ref{e11} are alternated epoch by epoch. As the training goes, the false-labeled samples will get greater loss while the true-labeled samples will get smaller loss with respect to the pre-generated label. Therefore, the false-labeled samples in $\mathcal{X}_{cl}$ and the true-labeled samples in $\mathcal{X}_{no}$ will be swapped in each alternation according to small-loss trick so as to progressively drive a better optimization.

To make the training process more stable, we utilize exponential-momentum-average (EMA) model $f_{ema}$ to self-generate pseudo label. It is inspired by \emph{Mean Teacher} \cite{Mean-teacher}, which has been widely used in many self-supervised and semi-supervised methods \cite{BYOL}. It helps generate more robust pseudo label via temporal ensemble. Therefore, the second term in Eqn.\ref{e11} can be reformulated as:
\begin{equation}
f_{ema}=\lambda f_{ema} + (1-\lambda)f
\label{e13}
\end{equation}
\begin{equation}
\min\limits_{f}\mathcal{L}_{ce}(f(\mathcal{X}_{no}^1), \arg\max  f_{ema}(\mathcal{X}_{no}^2))
\label{e14}
\end{equation}
The entire learning process is visualized in Fig.\ref{SSNLL}.

\subsection{Framework Overview}
\begin{algorithm}[htb]
   \caption{SSNLL for source-free UDA.}
   \label{alg}
\begin{algorithmic}
   \STATE {\bfseries Input:} Source domain pre-trained model $f$, unlabeled target data $\mathcal{X}_t$
   \STATE {\bfseries Output:} Optimized model $f$ 
   \STATE Update the statistics of BN layer via AdaBN (Eqn.\ref{e4})
   \STATE Pre-generate pseudo label $\mathcal{Y}_t$ (Eqn.\ref{e2})
   \STATE Refine the pseudo label $\widehat{\mathcal{Y}_t}$ via DTC  (Eqn.\ref{e5},\ref{e6},\ref{e7})
   \FOR{$e=1$ {\bfseries to} $epochs$}
   \STATE Split $\mathcal{X}_t$ into $\mathcal{X}_{cl}$ and $\mathcal{X}_{no}$ via small-loss trick (Eqn.\ref{e12})
   \STATE Fine-tune $f$ via training $\mathcal{X}_{cl}$ with pre-generated label and $\mathcal{X}_{no}$ with self-generated label (Eqn.\ref{e11},\ref{e13},\ref{e14})
   \ENDFOR
\end{algorithmic}
\end{algorithm}
To summarize, as shown in Algorithm.\ref{alg}, our framework is composed of three parts, including pseudo label pre-generating, label denoising preprocessing, and self-supervised noisy label learning. 

\section{Experiments}

\subsection{Datasets}
\paragraph{Digit datasets.}
We evaluate our method on three popular digit datasets, including: MNIST~\cite{mnist}, USPS~\cite{usps},  and Street View House Numbers (SVHN)~\cite{svhn}. 
MNIST contains $70K$ grayscale handwritten digit images with clean background ($60K$ for training and $10K$ for testing). USPS provides $9298$ grayscale handwritten digit images with unconstrained conditions ($7291$ for training and $2007$ for testing). SVHN includes cropped color digit images from a significantly harder, unsolved, real world scenes ($73K$ for training and $26K$ for testing). These digit datasets share 10 classes (0$\sim$9). However, the feature distributions are rather different, which is challenging for UDA evaluation. 
Concretely, following previous work \cite{2018Maximum,STAR}, we set the UDA tasks as MNIST$\rightarrow$USPS, USPS$\rightarrow$MNIST, SVHN$\rightarrow$MNIST. 

\paragraph{Traffic sign datasets.}
Our method is also evaluated on two traffic sign datasets: Synthetic Signs (Syn.Signs)~\cite{synsigns} and German Traffic Sign Recognition Benchmark (GTSRB)~\cite{GTSRB}. Both of them share 43 classes, where Syn.Signs contains $100K$ synthetic generated traffic sign images and GTSRB provides more than $50K$ traffic sign images collected from unconstrained natural scenes. Following previous work \cite{2018Maximum,STAR}, we set the UDA task as Syn.Signs$\rightarrow$GTSRB.

\paragraph{VisDA-C dataset.} 
\begin{figure}[tb]
\vskip 0.2in
\begin{center}
\centerline{\includegraphics[width=0.95\columnwidth]{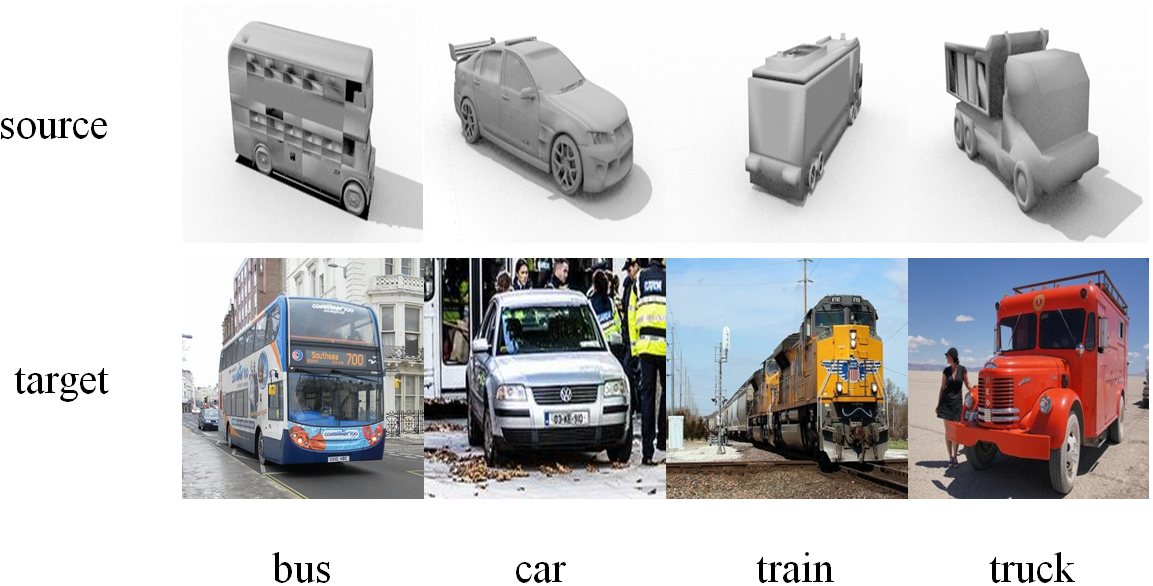}}
\caption{VisDA-C is a challenging large-scale object recognition dataset whose source data is synthesized by rendering 3D models and target data is sampled from real scenarios. Here we represent four challenging fine-grained classes on VisDA-C.}
\label{Dataset}
\end{center}
\vskip -0.2in
\end{figure}
VisDA-C~\cite{visda} is also a popular UDA evaluation benchmark, where the task is set as synthetic domain (train set) transferred to real domain (validation set) with 12 shared classes. The train set contains $152K$ synthetic images generated by rendering 3D models, and validation set consists of $55K$ real images sampled from a complex scene context dataset, namely MSCOCO~\cite{mscoco}. Since source domain contains large amount of data and target domain is extraordinary complex, this benchmark is one of the most challenging tasks for unsupervised domain adaptation, let alone source-free one. On the other hand, it can stress the superiority of our method without accessing source data during unsupervised adaptation.

\subsection{Experimental Settings}
\paragraph{Network architectures.}
In ours experiments, we strictly follow MCD~\cite{MCD} to design network architectures so as for fair performance comparison, which had also been adopted in many other related work~\cite{STAR,ADR,BSP,SHOT-1M}. Specifically, we adopt different modified LeNet networks for digit and traffic sign recognition tasks. One can refer to the code of MCD~\cite{MCD} in github\footnote{https://github.com/mil-tokyo/MCD\_DA/tree/master/classification}. As for VisDA-C, we directly use vanilla ResNet-101~\cite{resnet} pre-trained on ImageNet as the baseline model. 

\paragraph{Training hyper-parameters.}
We also follow MCD~\cite{MCD} to set the training hyper-parameters. Concretely speaking, for digit and traffic sign datasets, we utilize Adam to optimize all the networks with fixed learning rate $2e-4$, weight decay $1e-4$, batch size $128$, and training epoch $200$. For VisDA-C, we fine-tune the given pre-trained ResNet-101 by using mini-batch SGD with learning rate $1e-3$, momentum $0.9$, batch size $32$, training epoch $100$ as well as cosine learning rate decaying policy. Besides, split ratio $r$ is the only one hyper-parameter used in SSNLL. Without any specific statement, we set it as 0.2 in all experiments in a conservative way.

\paragraph{Implementation Details.}
In each iteration during training, we evenly sample images from $\mathcal{X}_{cl}$ and $\mathcal{X}_{no}$ to form each batch (1:1). To avoid regularizing the refined label in $\mathcal{X}_{no}$ to the most frequently appearing label in $\mathcal{X}_{cl}$, we use balanced sampling strategy to sample images of each class in $\mathcal{X}_{cl}$. To further mitigate the negative effects from noisy label and avoid over-fitting to the noisy label, an extra softmax layer is added at the top of the base network $f$ for prediction blurring during network training. 

\subsection{Experiments on Digit and Traffic Sign Recognition}
\begin{table*}[tb]
\caption{Performance on digit and traffic sign datasets compared with other state-of-the-art methods. Our result is reported by averaging five repetitions. \textcolor{red}{\textbf{Red}} bold font and \textcolor{blue}{\textbf{blue}} bold font denote the best results of source-free and source-based UDA methods, respectively. S2M, M2U, U2M, S2G are short for SVHN$\rightarrow$MNIST, MNIST$\rightarrow$USPS, USPS$\rightarrow$MNIST, and SynSigns$\rightarrow$GTSRB, respectively. '-' means the result is not reported in the original paper.}
\renewcommand\tabcolsep{5.0pt}
\label{experiments-digit}
\vskip 0.15in
\begin{center}
\begin{small}
\begin{sc}
\begin{tabular}{lccccc}
\toprule
 Methods & Source & S2M & M2U & U2M & S2G \\
\midrule
CDAN + E\cite{Long2017Conditional} &  $\surd$ & 89.2 & 95.6 & 98.0 & - \\
ADR \cite{ADR} & $\surd$ & 95.0$\pm$1.87 & 93.2$\pm$2.46 & 96.1$\pm$0.29 & - \\
MCD \cite{2018Maximum}& $\surd$ & 96.2$\pm$0.4 & 96.5$\pm$0.3 & 94.1$\pm$0.3 & 94.4$\pm$0.3 \\
CDAN + BSP \cite{BSP} & $\surd$ & 92.1 & 95.0 & 98.1 & - \\
rRevGrad + CAT \cite{2020Cluster} & $\surd$ & 98.8$\pm$0.0 & 94.0$\pm$0.7 & 96.0$\pm$0.9 & -\\
STAR \cite{STAR}& $\surd$ & 98.8$\pm$0.05 & 97.8$\pm$0.1 & \textcolor{blue}{\textbf{97.7$\pm$0.05}} & 95.8$\pm$0.2 \\
SWD \cite{Lee2020Sliced} & $\surd$ & \textcolor{blue}{\textbf{98.9$\pm$0.1}} & \textcolor{blue}{\textbf{98.1$\pm$0.1}} & 97.1$\pm$0.1 & \textcolor{blue}{\textbf{98.6$\pm$0.3}}\\
SHOT \cite{SHOT-1M}& $\times$ & 98.9$\pm$0.1 & \textcolor{red}{\textbf{97.9}$\pm$0.3} & 98.2$\pm$0.7 & -\\
\midrule
Source Only& $/$ & 74.9$\pm$0.3 & 48.1$\pm$3.0 & 29.7$\pm$5.0 & 86.5$\pm$3.1\\
Source Only + AdaBN& $/$ & 75.0$\pm$0.2 & 88.5$\pm$2.5 & 53.8$\pm$3.3 &84.8$\pm$2.3\\
Source Only + AdaBN + DTC& $/$ & 76.1$\pm$0.2 & 89.8$\pm$1.7 & 56.5$\pm$3.5 & 85.3$\pm$2.5\\
SSNLL (ours) & $\times$ & \textcolor{red}{\textbf{99.3}$\pm$0.05} & 97.1$\pm$0.1 & \textcolor{red}{\textbf{98.8}$\pm$0.1} & \textcolor{red}{\textbf{98.4}$\pm$0.1} \\
\bottomrule
\end{tabular}
\end{sc}
\end{small}
\end{center}
\vskip -0.1in
\end{table*}
\begin{figure}[tb]
\vskip 0.2in
\begin{center}
\centerline{\includegraphics[width=0.8\columnwidth]{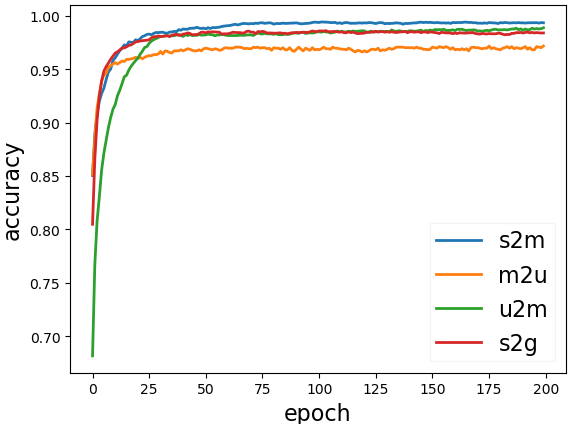}}
\caption{Comparison of our method among different adaptation tasks, including SVHN$\rightarrow$MNIST (s2m), MNIST$\rightarrow$USPS (m2u), USPS$\rightarrow$MNIST (u2m) and Syn.Signs$\rightarrow$GTSRB (s2g). We compute the accuracy on target data at each epoch.}
\label{digitresults}
\end{center}
\vskip -0.2in
\end{figure}
\begin{figure}[tb]
\vskip 0.2in
\begin{center}
\centerline{\includegraphics[width=0.8\columnwidth]{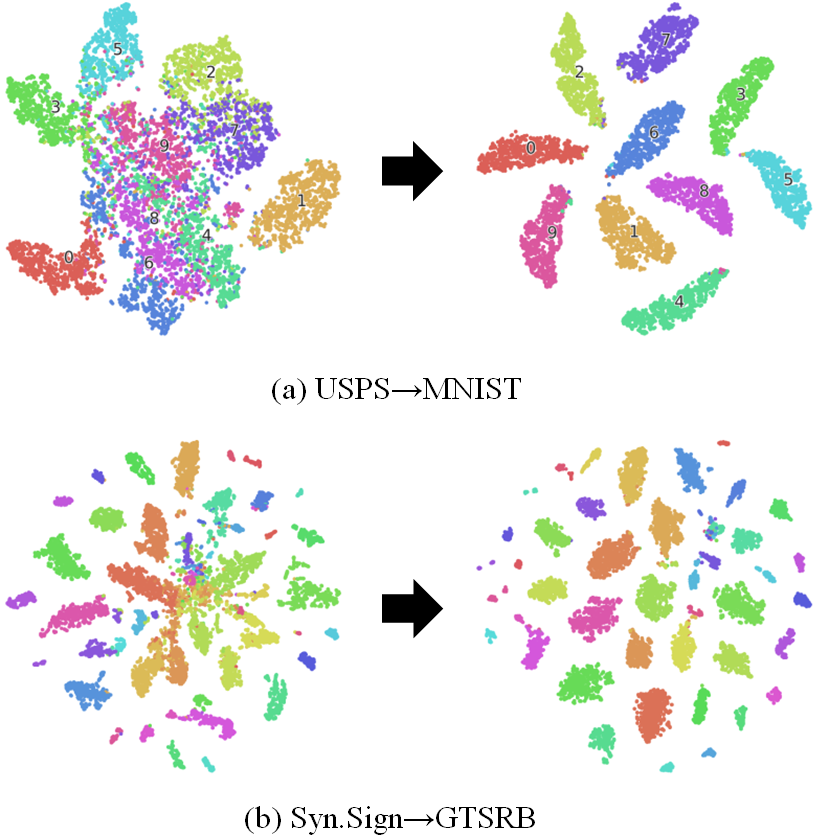}}
\caption{t-SNE visualization of the target features from the last second fully-connected layer. (a) is the USPS$\rightarrow$MNIST task before and after adaptation. (b) is the Syn.Sign$\rightarrow$GTSRB task before and after adaptation. Different colors denote different classes.}
\label{TSNE}
\end{center}
\vskip -0.2in
\end{figure}
In this section, we conduct four adaptation tasks to validate our approach, including SVHN$\rightarrow$MNIST, MNIST$\rightarrow$USPS, USPS$\rightarrow$MNIST, and Syn.Signs$\rightarrow$GTSRB. The detailed experimental results and performance comparison are shown in Tab.\ref{experiments-digit}. Our results are reported by averaging five repetitions with exactly the same settings under single-view inference. In the tasks of SVHN$\rightarrow$MNIST and USPS$\rightarrow$MNIST, our method surpasses other state-of-the-art methods by a large margin, no matter source-based or source-free UDA. In the task of MNIST$\rightarrow$USPS, the performance is a bit weaker, and we impute it to the reason that USPS is a small-scale dataset while self-supervised learning requires large-scale dataset. In the task of traffic sign adaptation task Syn.Signs$\rightarrow$GTSRB, our method surpasses most of state-of-the-art methods by a large margin and is comparable with SWD~\cite{Lee2020Sliced} which is a source data-based UDA method. In general, considering our method is a source-free one, we believe the advantage is very remarkable.

\paragraph{Label Denoising Preprocessing.} An important assumption in all source-free UDA methods is that the ratio of true-labeled samples should be larger than any other false-labeled classes. In this way, it is intuitive that to increase the initial ratio of true-labeled samples is beneficial for solving source-free UDA problem. As shown in Tab.\ref{experiments-digit}, we present the benefit of AdaBN and DTC for label denoising. We take USPS$\rightarrow$MNIST as an example for illustration. Before AdaBN and DTC, the accuracy is only 29.7\%, which is nearly randomly labeled and is extremely difficult for source-free unsupervised adaptation. After label denoising, the initial accuracy is improved to 53.8\% and 56.5\%, respectively, which makes the following source-free adaptation solvable.

\paragraph{Training Stability.} Accuracy is not the only one metric to evaluate the performance of UDA methods. How to achieve the results as steadily as supervised counterpart is much more important. We train four adaptation tasks for 200 epochs and present the accuracy curve $\rm{w.r.t}$ each epoch. As shown in Fig.\ref{digitresults}, the accuracy trends are improved steadily and achieve the best results as learning goes without performance degradation. This can reflect the effectiveness of our method from another aspect.

\paragraph{t-SNE Visualization.} Since our objective loss is composed of supervised and self-supervised classification loss, the features in target domain will get more discriminative after learning. As shown in Fig.\ref{TSNE}, we extract the features from the last second fully-connected layer for t-SNE visualization in the target domain of the tasks USPS$\rightarrow$MNIST and Syn.Signs$\rightarrow$GTSRB. The feature distribution in target domain is in a mess before adaptation. After our proposed SSNLL, the feautres with similar semantic information are well-grouped into one class tightly after adaptation.

\subsection{Experiments on Object Recognition}

\begin{table*}[tb]
\renewcommand\tabcolsep{3.0pt}
\caption{Performance on VisDA-C dataset compared with other state-of-the-art methods. \textcolor{red}{\textbf{Red}} bold font and \textcolor{blue}{\textbf{blue}} bold font denote the best results of source-free and source-based UDA methods, respectively. (Best viewed in color.)}
\label{experiments-visda}
\vskip 0.15in
\begin{center}
\begin{small}
\begin{sc}
\begin{tabular}{lcccccccccccccc}
\toprule
 Methods & \rotatebox{90}{Source}  &\rotatebox{90}{plane} & \rotatebox{90}{bcycl} & \rotatebox{90}{bus} & \rotatebox{90}{car} & \rotatebox{90}{horse} & \rotatebox{90}{knife} & \rotatebox{90}{mcycl} & \rotatebox{90}{person} & \rotatebox{90}{plant} & \rotatebox{90}{sktbrd} & \rotatebox{90}{train} & \rotatebox{90}{truck} &Avg\\
\midrule
 MCD \cite{2018Maximum}& $\surd$ & 87.0 & 60.9 & 83.7 & 64.0 & 88.9 & 79.6 & 84.7 & 76.9 & 88.6 & 40.3 & 83.0 & 25.8 & 71.9\\ 
ADR \cite{ADR}& $\surd$ & 94.2 &48.5 &84.0 &72.9 &90.1 &74.2 &92.6 &72.5 &80.8 &61.8 &82.2 &28.8 &73.5\\
CDAN \cite{Long2017Conditional}& $\surd$ & 85.2 &66.9 &83.0 &50.8 &84.2 &74.9 &88.1 &74.5 &83.4 &76.0 &81.9 &38.0 &73.9\\
CDAN + BSP \cite{BSP} & $\surd$ & 92.4 & 61.0 &81.0 &57.5 &89.0 &80.6 &90.1&77.0 &84.2 &77.9 &82.1 &38.4 &75.9\\
SAFN \cite{SAFN}& $\surd$ & 93.6 &61.3 &84.1 &70.6 &\textcolor{blue}{\textbf{94.1}} &79.0 &\textcolor{blue}{\textbf{91.8}} &\textcolor{blue}{\textbf{79.6}} &89.9 &55.6 &\textcolor{blue}{\textbf{89.0}} &24.4 &76.1\\
SWD \cite{Lee2020Sliced}& $\surd$ & 90.8 &82.5 &81.7 &70.5 &91.7 &69.5 &86.3 &77.5 &87.4 &63.6 &85.6 &29.2 &76.4\\
STAR \cite{STAR}& $\surd$ & \textcolor{blue}{\textbf{95.0}} & \textcolor{blue}{\textbf{84.0}} & \textcolor{blue}{\textbf{84.6}} & \textcolor{blue}{\textbf{73.0}} & 91.6 & \textcolor{blue}{\textbf{91.8}} & 85.9 & 78.4 & \textcolor{blue}{\textbf{94.4}} & \textcolor{blue}{\textbf{84.7}} & 87.0 & \textcolor{blue}{\textbf{42.2}} & \textcolor{blue}{\textbf{82.7}}\\
SHOT \cite{SHOT-1M}& $\times$ & 92.6 & 81.1 & 80.1 & 58.5 & 89.7 & 86.1 & 81.5 & 77.8 & 89.5 & 84.9 & 84.3 & \textcolor{red}{\textbf{49.3}} & 79.6\\
Model Adaptation \cite{MADA}& $\times$ & 94.8 & 73.4 & 68.8 & \textcolor{red}{\textbf{74.8}} & 93.1 & \textcolor{red}{\textbf{95.4}} & 88.6 & \textcolor{red}{\textbf{84.7}} & 89.1 & 84.7 & 83.5 & 48.1 & 81.6\\
\midrule
Source Only &/&85.0&6.0&61.2&84.1&49.4&0.1&71.8&4.1&61.2&29.5&62.1&0.7&42.9\\
Source Only + AdaBN &/&86.6&54.8&81.0&44.2&84.7&43.8&87.1&61.0&73.0&43.1&83.8&29.3&64.4\\
Source Only + AdaBN + DTC &/&92.7&58.9&79.3&52.9&88.2&49.8&85.9&70.7&79.6&58.9&82.2&31.2&69.2\\
SSNLL ($r=0.2$, ours) &$\times$ & \textcolor{red}{\textbf{98.1}} &86.2 & 89.1 & 74.2 & 95.6 & 89.0 & 92.5 & 76.4 & 94.1 & 88.3 & \textcolor{red}{\textbf{91.5}} & 47.1 & 85.2 \\
SSNLL ($r=0.4$, ours) &$\times$  & 97.2 & \textcolor{red}{\textbf{87.7}} & \textcolor{red}{\textbf{89.1}} & 73.6 & \textcolor{red}{\textbf{96.1}} & 91.2 & \textcolor{red}{\textbf{92.7}} & 79.9 & \textcolor{red}{\textbf{94.2}} & \textcolor{red}{\textbf{89.0}} & 90.4 & 48.9 & \textcolor{red}{\textbf{85.8}}\\
\bottomrule
\end{tabular}
\end{sc}
\end{small}
\end{center}
\vskip -0.1in
\end{table*}

In this section, we conduct experiments on VisDA-C~\cite{visda}, which is one of the most challenging datasets in UDA problems. The challenges lie in that the source data is synthesized by rendering 3D model while the target data is sampled in real scenarios. Also, some classes in this task share similar features, like bus, car, train, and truck, which transform it into a fine-grained classification problem. The experimental results and performance comparison are reported in Tab.\ref{experiments-visda}. Following \cite{MCD, STAR}, here we only report the class-wise accuracy under single-view inference. 

First of all, our two label denoising tricks, AdaBN and DTC, can improve the baseline accuracy from 42.9\% to 64.4\% and 69.2\%, respectively. Before label denoising preprocessing, the accuracy of some classes is even lower to 0.1\%, which is nearly impossible to optimize in an unsupervised way. After label denoising preprocessing and our proposed SSNLL, the accuracy can be improved to 85.8\%, which surpasses other state-of-the-art methods by a very large margin (more than 3.1\%), including those source data-based approaches. 

\paragraph{Performance on fine-grained classes.} We note that four fine-grained classes (bus, car, train, truck, as shown in Fig.\ref{Dataset}) on VisDA-C are quite easy to mix-classified into each other. Through carefully comparing the existing methods, as shown in Tab.\ref{experiments-visda2}, SHOT \cite{SHOT-1M} and Model Adaptation \cite{MADA} perform not well on bus and car, respectively. SAFN \cite{SAFN}, SWD \cite{Lee2020Sliced} and STAR \cite{STAR} perform not well on truck. We find that our method can well-improve the accuracy of these fine-grained classes and perform the best among the existing methods.

\begin{figure}[tb]
\vskip 0.2in
\begin{center}
\centerline{\includegraphics[width=0.8\columnwidth]{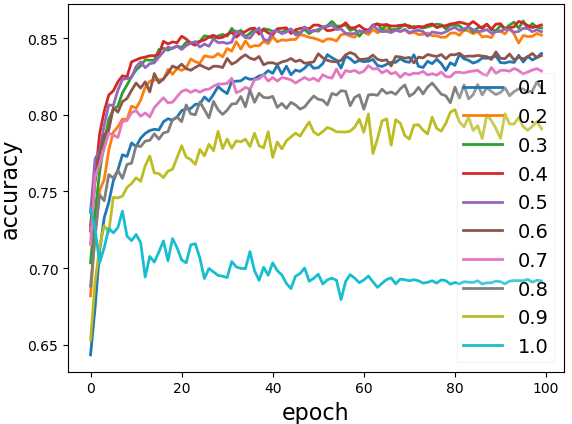}}
\caption{The ablation study of the hyper-parameter \emph{split ratio} $r$ on VisDA-C varying from 0.1 to 1.0, where 1.0 means that we directly use all samples with noisy label for fune-tuning. We compute the accuracy on the validation set of VisDA-C at each epoch.}
\label{splitratio}
\end{center}
\vskip -0.2in
\end{figure}
\begin{table}[tb]
\renewcommand\tabcolsep{3.0pt}
\caption{Performance comparison of four challenging fine-grained classes on VisDA-C. \textcolor{red}{\textbf{Red}} bold font and \textcolor{blue}{\textbf{blue}} bold font denote the best results of source-free and source-based UDA methods, respectively. The accuracy on this table is extracted from Tab.\ref{experiments-visda}.}
\label{experiments-visda2}
\vskip 0.15in
\begin{center}
\begin{small}
\begin{sc}
\begin{tabular}{lcccccc}
\toprule
 Methods & \rotatebox{90}{Source}  & \rotatebox{90}{bus} & \rotatebox{90}{car} & \rotatebox{90}{train} & \rotatebox{90}{truck} &Avg\\
\midrule
SAFN & $\surd$ & 84.1 &70.6 &\textcolor{blue}{\textbf{89.0}} &24.4 &67.0\\
SWD& $\surd$ & 81.7 &70.5 &85.6 &29.2 &66.8\\
STAR& $\surd$ & \textcolor{blue}{\textbf{84.6}} & \textcolor{blue}{\textbf{73.0}} & 87.0 & \textcolor{blue}{\textbf{42.2}} & \textcolor{blue}{\textbf{71.7}}\\
SHOT& $\times$ & 80.1 & 58.5 & 84.3 & \textcolor{red}{\textbf{49.3}} & 68.1\\
Model Adaptation & $\times$ & 68.8 & \textcolor{red}{\textbf{74.8}} & 83.5 & 48.1 & 68.8\\
\midrule
SSNLL ($r=0.2$, ours) &$\times$ & 89.1 & 74.2 & \textcolor{red}{\textbf{91.5}} & 47.1 & 75.4 \\
SSNLL ($r=0.4$, ours) &$\times$ & \textcolor{red}{\textbf{89.1}} & 73.6 & 90.4 & 48.9 & \textcolor{red}{\textbf{75.5}}\\
\bottomrule
\end{tabular}
\end{sc}
\end{small}
\end{center}
\vskip -0.1in
\end{table}

\paragraph{Ablation study on split ratio $r$.} Split ratio $r$ is an important hyper-parameter in SSNLL. Intuitively, $r$ is strongly related to the noisy ratio of the pre-generated pseudo label. Unfortunately, the noisy ratio is unknown in practical scenarios. Therefore, we mainly set $r$ as a small value 0.2 in a conservative way. In the experiments on VisDA-C, we find $r=0.4$ can achieve the best results. To thoroughly analyze the influence of $r$, we carry out extensive experiments on VisDA-C with $r$ varying from 0.1 to 1.0. As shown in Fig.\ref{splitratio}, only $r=1.0$ and $r=0.9$ perform the worse. The accuracy of other settings are all greater than 80.0\%. Specifically, the accuracy of $r=0.2\sim 0.5$ is greater than 85.0\%. In general, it is safe to set $r$ as a small value.

\section{Conclusion}
In this paper, we model source-free unsupervised domain adaptation problem into learning from noisy label. From this perspective, we propose a Self-Supervised Noisy Label Learning method, which is mainly composed of two critical steps. One is to split the target data into a cleaner subset and a noisier subset via small-loss trick. Another is to uniformly sample the image from these two subsets equipped with pre-generated label and self-generated label to fine-tune the given source domain pre-trained model. The former one regularizes the latter one to refine their self-generated label. These two steps are alternated epoch by epoch to progressively boost the performance. Our method surpasses other methods even source data-based methods by a very large margin. We hope our approach can bring inspirations for the UDA community.

\bibliography{example_paper}
\bibliographystyle{icml2021}

\end{document}